\documentclass[a4paper,fleqn,elsarticle]{cas-sc}
\usepackage[authoryear,longnamesfirst]{natbib}
\usepackage{amsmath}

\usepackage[export]{adjustbox}
\usepackage{standalone}
\usepackage{tikz}
\usetikzlibrary{patterns}
\usetikzlibrary{shapes.geometric}
\usetikzlibrary{positioning}
\usetikzlibrary{arrows.meta}
\usetikzlibrary{fit}
\usetikzlibrary{backgrounds}
\usetikzlibrary{calc}
\usepackage{pgfplots}
\usepackage{pgfplotstable}
\usepackage{tabularx}
\usepackage{algorithm2e}
\usepackage{float}
\usepackage{placeins}
\usepackage{booktabs}
\usepackage{multirow}
\usepackage{colortbl}
\usepackage{makecell}
\usepackage[T1]{fontenc}
\usepackage{caption}

\usepackage{xcolor}
\definecolor{revblue}{RGB}{0,0,200}

\definecolor{provorange}{RGB}{200,90,0}
\newcommand{\prov}[1]{#1}
\newcommand{\vnext}[1]{\emph{[Deferred to the corpus-scale version: #1]}}

\newdefinition{rmk}{Remark}

\newcommand{\prismaDBqueries}{6}
\newcommand{\prismaDBrecords}{45}
\newcommand{\prismaDBdups}{2}
\newcommand{\prismaDBunique}{43}
\newcommand{\prismaCiteIdentified}{10056}

\newcommand{\prismaReportsRetrieved}{286}
\newcommand{\prismaReportsExcluded}{48}
\newcommand{\prismaExclNotEntitled}{10}
\newcommand{\prismaExclNoFormulas}{15}
\newcommand{\prismaExclTierThree}{23}
\newcommand{\prismaInclPapers}{238}
\newcommand{\prismaInclFormulations}{8957}

\ExplSyntaxOn
\cs_gset:Npn \__first_footerline:
{
  \group_begin:
  \small
  \sffamily
  \ifnum\theblind>0\relax
  \else
	\__short_authors: :~
  \fi
  { \rmfamily \itshape Preprint }
  \group_end:
}
\ExplSyntaxOff

\begin{document}
\let\WriteBookmarks\relax
\def\floatpagepagefraction{1}
\def\textpagefraction{.001}

%% --- Front matter ---
\shorttitle{LP Mining with LP2Graph}
\shortauthors{J. Maurischat et al.}
\title[mode = title]{LP Mining with LP2Graph: A Use Case for Railway Rescheduling}

\author[1]{J\"orn Maurischat}[ orcid=0009-0005-6050-7596 ]
\cormark[1]
\ead{joern.maurischat@tu-dresden.de}
\credit{Conceptualization, Methodology, Software, Data curation, Investigation, Visualization, Writing -- original draft}
\author[1]{Nikola Be\v{s}inovi\'c}
\credit{Conceptualization, Supervision, Writing -- review \& editing}
\author[2]{Michael F\"arber}
\credit{Supervision, Writing -- review \& editing}

\affiliation[1]{organization={TU Dresden},
  addressline={Chair of Railway Operations,"Friedrich List" Faculty of Transport and Traffic Sciences},
  city={Dresden}, postcode={01069}, country={Germany}}
\affiliation[2]{organization={TU Dresden},
  addressline={ScaDS.AI Dresden/Leipzig, Chair of Scalable Software Architectures for Data Analytics},
  city={Dresden}, postcode={01069}, country={Germany}}
\cortext[cor1]{Corresponding author}

%% Abstract lives in its own file (chapters/00_frontmatter/abstract.tex), which
%% contains the whole abstract environment. A NORMAL \input here works; do not
%% \input *inside* the abstract env (cas-sc writes the body verbatim to .abs and
%% mangles a nested \input / a lone macro). See abstract.tex header for details.
%% Abstract --- single source of truth, in its own file.
%% Main.tex pulls it in with a NORMAL \input at the front-matter point
%% (\input{chapters/00_frontmatter/abstract}). The whole abstract environment
%% lives here so the literal text is what cas-sc writes verbatim to \jobname.abs.
%% (Do NOT \input a file *inside* the abstract env, and do NOT wrap the text in a
%% macro --- cas-sc's verbatimwrite drops the \input token / eats a lone macro,
%% leaving an empty or broken abstract. Inputting the whole env from here works.)
\begin{abstract}
Like many optimization-driven domains, railway rescheduling relies on Mixed-Integer Linear Programming (MILP), yet the field's modeling knowledge is scattered across hundreds of papers in incompatible notations, and narrative surveys organize it subjectively: they classify models by vocabulary rather than by structure, and reproduce neither. We present \emph{LP Mining with LP2Graph}, a method that mines the structure of published LP and MILP formulations into a reproducible dataset and an induced taxonomy. Its core, \emph{LP2Graph}, represents each formulation admitted by its canonical grammar as a typed variable--equation graph derived from a single canonical model; once a source is extracted into that model, everything downstream is deterministic. Each source is parsed into this model, homologized, and clustered bottom-up (over variables, then constraints and the objective, then whole-model structure) and, separately, by application domain and solution approach; the resulting groups are labeled by a rule-seeded, self-updating classifier. We validate the representation rather than assume it: per-cluster representatives are regenerated as independent \LaTeX{} and re-solved across CBC, HiGHS and Gurobi against the optimum reported in the source paper. The outcome is an objective, repeatable taxonomy of variables, constraints and model types: the principled foundation on which our \emph{raiLPminer} line of automated railway-rescheduling model development builds.
\end{abstract}

\begin{highlights}
\item A deterministic, graph-based method for mining the structure of LP and MILP formulations from the literature
\item LP2Graph typed variable--equation graphs enable solver-free structural comparison and feature extraction
\item Bottom-up multi-level clustering and two-stage labeling yield a reproducible taxonomy of variables, constraints and model types
\item Representation fidelity validated by round-trip translation and cross-solver reproduction of published optima
\end{highlights}

\begin{keywords}
mixed integer linear programming \sep
railway rescheduling \sep
optimization model mining \sep
graph representation \sep
taxonomy \sep
reproducible literature analysis \sep
raiLPminer
\end{keywords}

\maketitle

%% --- Chapters ---
\section{Introduction} \label{sec:1}

Many domains, such as production scheduling, vehicle routing, service-network
design and energy dispatch, cast their hardest planning and control decisions as
Mixed-Integer Linear Programs (MILP) \citep{nemhauser1988}, and each has deposited
decades of operations-research (OR) modeling knowledge in published formulations.
This paper is about making that knowledge machine-readable; railway rescheduling
is the application area on which we demonstrate it, and a demanding one. When
disruptions invalidate a planned timetable, operators must reroute, reorder and
retime trains under tight safety, capacity and connection constraints, and OR
research has produced a rich catalogue of recovery models and algorithms for
doing so \citep{cacchiani2014, ReschReview2023, besinovic2020, railrev2022}. This
catalogue, however, lives almost entirely as prose and mathematics scattered
across hundreds of papers, each written in its own notation. The modeling knowledge that matters (which variables carry which
decisions, which constraint families recur, how whole formulations are
structured) is dispersed and, in its current form, neither searchable nor
comparable across sources.

% --- GAP 1: narrative reviews are subjective, non-reproducible, non-structural.
The conventional way to consolidate such a body of work is a narrative literature
review or survey. These summarize the field subjectively: they can be neither
reproduced nor quantified, and they classify formulations by domain vocabulary
rather than by the structure of the formulations themselves
\citep{ReschReview2023, railrev2022}. Two formulations that share the same
underlying structure but use different words are filed apart; two that share words
but differ structurally are filed together. The consequence is concrete: a
practitioner who must re-order and re-time trains through a disrupted bottleneck
and reaches for a published model will find a railway timetable-rescheduling
formulation \citep{veelenturf2016rescheduling} and a job-shop scheduling
formulation \citep{ku2016jobshop} filed in different literatures. Yet, read
structurally, both are disjunctive ``who-goes-first'' ordering models, and for
microscopic bottleneck dispatching the job-shop model is arguably the closer fit,
train dispatching being job-shop scheduling with blocking
\citep{dariano2008reordering}. The title page names the domain; only the
structure tells whether a model fits the problem, and a vocabulary-filed survey
cannot surface that. There is, today, no objective,
reproducible, structural account of how railway-rescheduling models are actually
built, and reviewers increasingly expect the selection process itself to follow
a documented systematic-review protocol such as PRISMA \citep{page2021prisma}
, which narrative reviews of this kind do not provide.

% --- GAP 2: railway rescheduling is structurally embedded in a wider
% transportation (and, analogically, production) family; that structural kinship
% is what licenses the two-axis, shell-prioritized corpus of Section~\ref{sec:scope_corpus}.
This deficiency is not confined to railways, and that structural kinship is what the
method exploits rather than a separate gap it claims to close. Timetable-recovery
formulations are close structural relatives of the wider transportation literature
and, more loosely, of production scheduling and of routing and scheduling models
generally, such as vehicle routing, job-shop scheduling and service-network design
\citep{toth2014vrp,pinedo2016scheduling,crainic2000snd},
with which they share variable patterns (assignment, sequencing, flow) and
constraint families (precedence, capacity, time windows). This kinship is why the
corpus is not restricted to railway rescheduling but ordered by structural
proximity to it: railway first, the broader transportation domain next, and
production as an analogical outer shell drawn on only where nearer domains are
sparse, spanning both the reactive rescheduling task and the wider set of planning
and control operations (Section~\ref{sec:scope_corpus}). Railway rescheduling is the
demonstration target; the comparable structure the method induces is what makes
that widening principled rather than opportunistic.

% --- GAP 3: solver code is rarely published and there is no deterministic
% text-to-structure path, so most modeling knowledge stays locked in prose/\LaTeX{}
% (cf. extraction, Section~\ref{sec:extraction}; source-artifact tiers, Section~\ref{sec:scope_corpus}).
A third gap concerns extraction and reproducibility. In this field it remains
uncommon to publish the solver code behind a formulation, which is poor practice
for reproducible research: a paper bound to a released model can be cloned and
re-run deterministically, mined exactly and validated in seconds, whereas a
PDF-only formulation must be recovered from its text. Reading the structure of a
model out of free text or \LaTeX{} (inspecting the formulation rather than
re-implementing it) has no fully deterministic path, leaving the bulk of the
literature's modeling knowledge locked in a form that cannot be processed at
scale.

% --- IDEA: LP Mining with LP2Graph; dataset = headline; comparability.
We address these gaps with \emph{LP Mining with LP2Graph}, presented as a use case
for a systematic, PRISMA-style literature review of railway rescheduling. At its
core, \emph{LP2Graph} is a deterministic procedure that represents a defined class
of LP and MILP formulations, namely those expressible in its canonical grammar
(Section~\ref{sec:lp2graph}), as typed variable--equation graphs derived from a
single canonical model. This paper is therefore \emph{not} a contribution to
LLM-based optimization modeling: no language model participates in the
representation, the mining, or the validation. Where recent work synthesizes new
models from natural-language descriptions, we do the converse: we mine the
\emph{published} literature into a structured, comparable, validated corpus; the
relation to automated model synthesis is deferred to the outlook
(Section~\ref{sec:outputs}). Around the representation we build a mining pipeline that takes a
prioritized corpus of published formulations, extracts and homologizes each into
this representation, derives lexical and structural feature vectors, clusters the
formulations bottom-up without prior names, and labels the resulting groups
top-down from existing taxonomy vocabulary. The induced clusters are then
sanity-checked against the classifications in existing expert review papers, and
the divergence between machine-induced structure and expert taxonomy is quantified
and explained, with those expert classifications serving as an interpretive anchor
rather than as ground truth. The headline output
is a \emph{dataset}: published models brought into one common form, labeled and
grouped, delivering structural \emph{comparability} across formulations that did
not exist before. We deliberately make the \emph{structure} comparable, not the
results: different objectives and data make ranking formulations meaningless, so we
normalize the foundations, not the outcomes.

% --- CONTRIBUTIONS (prose, enumerated inline per house style).
This paper makes four contributions. (i) \emph{LP2Graph}, a deterministic typed
variable--equation graph representation of LP/MILP formulations with views,
structural metrics and bidirectional codecs between the graph and concrete model
sources: \LaTeX{}$\leftrightarrow$graph, currently the primary channel, and
code$\leftrightarrow$graph, importing formulations from solver code and exporting
executable models back out. (ii) An end-to-end
\emph{mining pipeline} (corpus construction, extraction/homologization, feature
construction, multi-level clustering and two-stage labeling) that operates on
the published literature. (iii) The mined \emph{dataset}, the primary
contribution, which establishes structural comparability across otherwise
incompatible formulations. (iv) An induced \emph{taxonomy} of variable roles,
constraint families and model types, anchored against and contrasted with expert
classifications. Re-solving a representative formulation per cluster serves only to
validate the \LaTeX{}$\to$graph$\to$translate$\to$solve round-trip, not to rank
papers.

The remainder of the paper is organized as follows. Section~\ref{sec:2} reviews the
relevant literature and states the gap precisely. Section~\ref{sec:3} details the
mining method and the LP2Graph representation. Section~\ref{sec:Experiments} and
Section~\ref{sec:result} describe the corpus and present the resulting dataset and
taxonomy. Section~\ref{sec:concl} concludes.

\section{Related Work}\label{sec:2}

% Structure modeled on Paper 2 (raiLParchitect) but restructured for Paper 1's
% mining frame, per JM 2026-06-21: less focus on per-element model detail; more
% on (a) extraction of models from text/code, (b) graph representations,
% (c) larger routing/scheduling problems, (d) reading structure out of text, and
% (e) strategic cross-domain transfer. Each subsection is prose and ends toward
% the gap; the section closes with an explicit gap synthesis.

\subsection{Railway rescheduling and the wider routing--scheduling family}
Railway timetable recovery has a deep MILP literature: rerouting, reordering and
retiming under capacity, headway and connection constraints, solved exactly or
with dedicated heuristics \citep{cacchiani2014, ReschReview2023, besinovic2020,
railrev2022}. Structurally, these formulations are close relatives of a much
larger body of routing and scheduling models, such as vehicle routing, job-shop
scheduling and service-network design \citep{toth2014vrp,pinedo2016scheduling,crainic2000snd},
which share variable patterns (assignment, sequencing, flow) and constraint
families (precedence, capacity, time windows) with timetable recovery. The
present work treats railway rescheduling as the demonstration domain while
recognizing that the modeling structures it targets recur across this family;
that proximity is what later licenses a strategic widening of the corpus
(Section~\ref{sec:3}).

\subsection{Surveys, taxonomies and systematic-review methodology}
The standard way to organize this literature is the narrative survey, which
classifies models by domain vocabulary and summarizes them subjectively
\citep{ReschReview2023, railrev2022}. Existing taxonomies of optimization models
do capture recurring variable, constraint and model types, but they are
expert-authored, not reproducible, and rest on naming rather than on the
structure of the formulations. Reviewers now increasingly expect review-shaped
papers to follow a documented systematic-review protocol, PRISMA
\citep{page2021prisma}, which specifies how a corpus is searched, screened and
reported. We position
this paper as a \emph{use case} for such a systematic review: a PRISMA-conformant
corpus protocol feeds a method that induces a taxonomy mechanically, and existing
expert classifications serve as an anchor to sanity-check the induced clusters
against rather than as the ground truth.

\subsection{Extracting formal models from papers and code}
A growing line of work extracts formal optimization models from natural-language
descriptions and turns them into solver code, much of it now LLM-driven
\citep{Optimus2024, ORLM2025, OptGen2022, SurveyORLM2024, ahmaditeshnizi2024,
wasserkrug2024, Formalization2025, li2023optiguide}, accompanied by dedicated benchmarks for
natural-language optimization modeling such as NL4Opt \citep{ramamonjison2023nl4opt},
NLP4LP \citep{ahmaditeshnizi2024} and MAMO \citep{huang2024mamo}. Within this line,
validation of the generated models is increasingly treated as a first-class concern:
TriVAL \citep{fang2026trival} inserts an explicit construct--validate--revise step at
each modeling stage (semantic specification, mathematical formulation, code), a
validation need to which the structural substrate mined here is directly complementary.
This effort is overwhelmingly oriented toward
\emph{synthesis}, producing one new model from a natural-language prompt, and
is evaluated on solve accuracy against curated problem descriptions. The present
work is neither a synthesis system nor a benchmark for one: it mines the
\emph{published} literature into a structured, comparable corpus, uses no language
model in its pipeline, and is therefore not comparable to these systems on their
own terms. The most
faithful source of an existing model is its released code: a paper bound to a
public repository can be mined deterministically and validated quickly, whereas
PDF-only formulations must be recovered from text. Reading a model's structure out
of its \LaTeX{} or prose (inspecting the formulation instead of
re-implementing it) is the data-extraction problem at the heart of this paper,
and it has no fully deterministic solution for unstructured sources.

\subsection{Graph representations of optimization models}
A MILP has a natural bipartite graph structure, with variables linked to the
equations in which they appear, and this representation is well established as a
basis for structural analysis and machine learning over combinatorial problems
\citep{gasse2019exact,bengio2021mlco}. Graph-structured views also support deterministic structural metrics (for
example size, constraint-to-variable ratio and graph diameter) that summarize a
formulation independently of its wording (cf.\ \citealp{diameterMetro,diameterStreet}). Prior uses of such graphs target learning to solve or to branch; here the same
bipartite structure is repurposed as a canonical, comparable \emph{representation}
for mining and clustering published models, so that structurally similar
formulations group together regardless of the vocabulary their authors chose.

\subsection{Cross-domain structural transfer and the shell-prioritized corpus}
Bringing heterogeneous formal models onto a common footing is not unique to
transport: structural-comparison and instance-synthesis efforts recur across the
wider optimization literature \citep{smithmiles2012instance,gleixner2021miplib}. This is what makes a structurally widening corpus viable. When railway
rescheduling alone is too sparse to support reliable clustering, structurally
analogous formulations from the broader transportation domain and, in the outer
analogical shell, production scheduling can be drawn in without breaking
comparability. The corpus is ordered along two axes: application domain by
proximity to railway, and activity from rescheduling out to general operations.
It therefore grows along a principled priority (cells $P_1$--$P_5$,
Section~\ref{sec:scope_corpus}) rather than opportunistically.

\subsection{Reproducibility and validation in OR}
Finally, reproducing a published optimization result is itself non-trivial when
neither data nor code is released, and a mined representation must be shown to be
faithful rather than assumed so. Round-trip translation (regenerating an
independent formulation from the graph), cross-solver re-solving and reproduction
of a paper's reported optimum are the available levers \citep{koch2011miplib2010}. In this work validation is applied a posteriori and only to confirm the
representation's round-trip fidelity, not to rank papers or claim one model
superior to another.

\subsection{Gap}
Across all of the above, the same gap recurs and is the one this paper targets.
There is no objective, reproducible, \emph{structural} account of how
railway-rescheduling (and, more broadly, routing and scheduling)
optimization models are built. Narrative surveys are subjective and
non-reproducible; existing taxonomies classify by vocabulary, not structure;
extraction work synthesizes new models instead of mining existing ones into a
comparable form; graph representations are used to solve rather than to compare;
and the bulk of the literature's modeling knowledge stays locked in text because
reading structure out of unstructured sources has no deterministic path. The
consequence is an absence of \emph{comparability}: no shared, labeled dataset of
variable roles, constraint families and model types against which formulations can
be placed side by side. \emph{LP Mining with LP2Graph} is designed to close
exactly this gap: it mines \emph{existing} models rather than synthesizing new
ones, uses a deterministic \emph{structural} representation, delivers a comparable
labeled dataset carrying an \emph{induced} taxonomy, and checks representation
fidelity \emph{a~posteriori}.

\section{Methodology}\label{sec:3}

This section develops a method for \emph{mining} the structure and content of published LP and MILP formulations and distilling them into a reproducible dataset and a taxonomy. The analysis is constructive, and its determinism boundary is stated precisely: extracting a formulation from a heterogeneous source into the canonical model involves interpretation and is audited rather than claimed exact (Section~\ref{sec:extraction}); \emph{from the validated canonical model onward}, the pipeline is deterministic (given the same corpus, configuration and versioned resources it produces the same feature vectors, clusters and rule-assigned labels), and the human-in-the-loop steps (ambiguous extractions and adjudicated labels) are logged so the end-to-end result is reproducible by replay rather than by re-judgment. The single design principle underlying every step is a \emph{cluster-and-name operator} (Section~\ref{sec:feature_clustering}): meaningful textual tokens are extracted, grouped into concepts through an acknowledged lexical resource, projected into a fixed-dimensional vector, clustered, and named after the most frequent formulation inside each cluster. The contribution of this paper is to apply that operator, together with the deterministic structural representation provided by \emph{LP2Graph}, in a disciplined bottom-up sweep over the constituent parts of an optimization model.

Figure~\ref{fig:mining_pipeline} summarizes the pipeline. A corpus of formulations is assembled and prioritized by proximity to the target problem (Section~\ref{sec:scope_corpus}). Each formulation is parsed from solver code, publisher full-text markup or \LaTeX{} into the LP2Graph canonical model and homologized (Section~\ref{sec:lp2graph}, Section~\ref{sec:extraction}). Lexical features over names and deterministic structural features over the graph are combined into feature vectors, which drive the multi-level clustering and naming (Section~\ref{sec:feature_clustering}): variables and parameters first, then constraints and the objective conditioned on the variables they couple, then whole-model structure; in parallel a text-only analysis clusters each model along the two dimensions of application domain and solution approach. The resulting groups receive multi-level, multi-dimensional labels from a two-stage mechanism that seeds labels with transparent rules, generalizes them with a supervised classifier, and writes confident predictions back into a versioned label database (Section~\ref{sec:labeling}). The faithfulness of the underlying representation is established empirically, not assumed: for the highest-cited representative of each cluster, LP2Graph regenerates the model, translates it into a second modeling language, re-solves it across independent solvers, and reproduces the optimum reported in the source paper (Section~\ref{sec:validation}). The outputs are a reproducible dataset and a new taxonomy (Section~\ref{sec:outputs}).

\begin{center}
\captionsetup{type=figure}
\begin{tikzpicture}[
  node distance = 5mm,
  font = \footnotesize,
  stage/.style = {rectangle, rounded corners=2pt, draw=black!70, fill=black!4,
                  text width=0.76\linewidth, align=center, inner sep=4pt, minimum height=7mm},
  val/.style   = {rectangle, rounded corners=2pt, draw=revblue!70, fill=revblue!4,
                  text width=0.76\linewidth, align=center, inner sep=4pt},
  flow/.style  = {-{Latex[length=2mm]}, thick, draw=black!70},
  back/.style  = {-{Latex[length=2mm]}, thick, draw=revblue!70, dashed},
]
\node[stage] (corpus) {\textbf{1. Corpus construction}\\ shell-prioritized LP/MILP sources (solver code, publisher XML/MathML, \LaTeX{})};
\node[stage, below=of corpus] (extract) {\textbf{2. Extraction \& homologization}\\ parse to LP2Graph canonical model; normalize symbols};
\node[stage, below=of extract] (feat) {\textbf{3. Feature construction}\\ lexical concepts $\oplus$ deterministic structural metrics $\rightarrow$ vectors};
\node[stage, below=of feat] (cluster) {\textbf{4. Multi-level clustering \& naming}\\ variables $\rightarrow$ constraints/objective $\rightarrow$ model type; text-only domain $\times$ solution approach};
\node[stage, below=of cluster] (label) {\textbf{5. Two-stage labeling}\\ rule seed $\rightarrow$ supervised classifier; closed-loop label database};
\node[stage, below=of label] (out) {\textbf{6. Outputs}\\ reproducible dataset $+$ taxonomy (variables, constraints, model types)};
\node[val, below=of out] (valid) {\textbf{Representation-fidelity validation}\\ graph $\rightarrow$ \LaTeX{} $\rightarrow$ graph round-trip; translate to a 2nd modeling language; re-solve (CBC/HiGHS/Gurobi) vs.\ published optimum};

\draw[flow] (corpus) -- (extract);
\draw[flow] (extract) -- (feat);
\draw[flow] (feat) -- (cluster);
\draw[flow] (cluster) -- (label);
\draw[flow] (label) -- (out);
\draw[back] (out) -- (valid);
\draw[back] (valid.east) -- ++(3mm,0) |- (extract.east);
\end{tikzpicture}

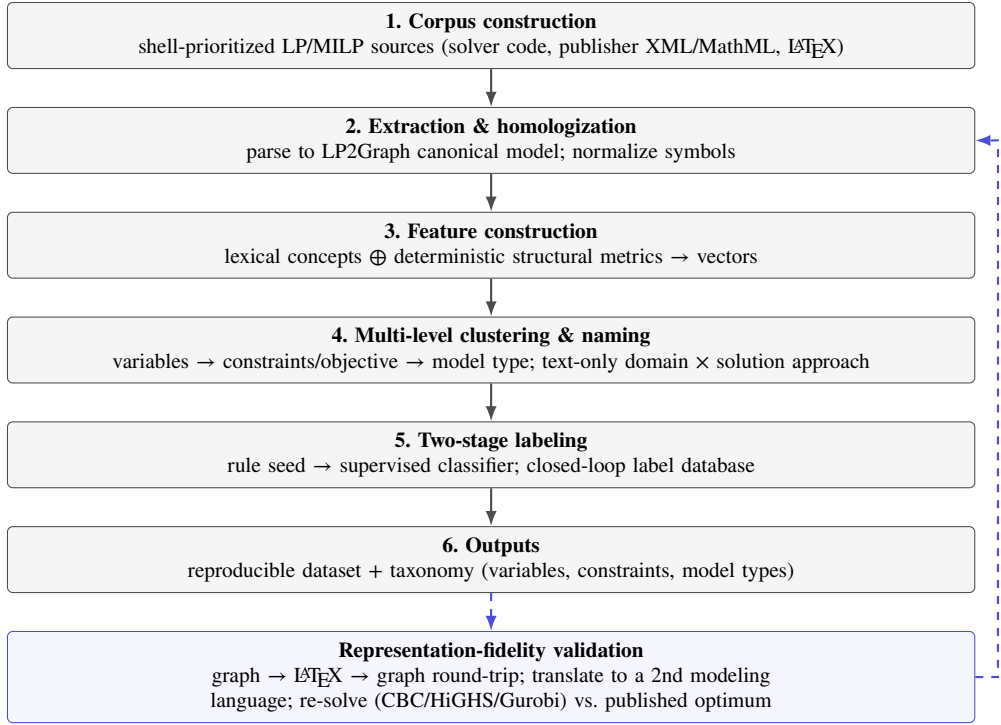
\captionof{figure}{The LP-mining pipeline. Solid arrows are the forward path from corpus to taxonomy, deterministic from the validated canonical model (stage~2 output) onward, with the extraction into that model audited separately (Section~\ref{sec:extraction}); the dashed loop is the \emph{a posteriori} representation-fidelity check applied to the highest-cited representative of each cluster.}
\label{fig:mining_pipeline}
\end{center}

\subsection{Scope and Corpus Construction}\label{sec:scope_corpus}

\subsubsection{Domain scope: a shell model}

The corpus is defined over two orthogonal axes. The first is the \emph{application domain}, ordered by structural proximity to the target problem: railway is the inner shell, the broader field of transportation is the middle shell, and production (manufacturing) forms an outer shell. These shells are not strict set inclusions (production scheduling is not a superset of transportation) but a deliberate ordering of \emph{structural transferability}: the further from railway, the looser the expected match between a source formulation and railway rescheduling, hence the lower its priority for inclusion. The second axis is the \emph{activity}: rescheduling, i.e.\ the reactive re-optimization of plans under disruption, which is the target task, is a proper subset of the wider set of operations (planning, control and dispatching). Figure~\ref{fig:domain_shells} visualizes both axes.

Crossing the two axes yields the inclusion priority. The four cells spanned by the in-scope transportation super-domain are ranked activity-first and domain-second: railway rescheduling ($P_1$), transportation rescheduling ($P_2$), railway operations ($P_3$) and transportation operations ($P_4$). Production rescheduling is appended as an outer analogical shell ($P_5$) and is drawn on only to augment clusters left sparse by the four in-scope cells; production operations is out of scope. Rescheduling is prioritized over general operations because disruption-response models carry the temporal-recovery and re-allocation structure most relevant to the target, while the domain ordering reflects decreasing notational and structural compatibility.

\begin{figure}[!t]
\centering
\begin{tikzpicture}[font=\scriptsize]
  % concentric domain shells (draw largest first so inner shells sit on top)
  \draw[fill=black!5,  draw=black!55, thick] (0,0) circle (3.9);
  \draw[fill=black!9,  draw=black!55, thick] (0,0) circle (2.7);
  \draw[fill=black!15, draw=black!55, thick] (0,0) circle (1.5);
  % activity divider
  \draw[dashed, draw=black!60] (-3.9,0) -- (3.9,0);
  % activity axis labels
  \node[align=center, font=\scriptsize\bfseries] at (0, 4.35) {RESCHEDULING\\[-1pt]\footnotesize\mdseries(reactive, under disruption)};
  \node[align=center, font=\scriptsize\bfseries] at (0,-4.35) {OPERATIONS\\[-1pt]\footnotesize\mdseries(planning \& control)};
  % cell labels (top = rescheduling, bottom = operations)
  \node[align=center] at (0, 0.72) {Railway\\rescheduling\\[1pt]\textbf{($P_1$)}};
  \node[align=center] at (0,-0.78) {Railway\\operations\\[1pt]\textbf{($P_3$)}};
  \node[align=center] at (0, 2.05) {Transportation\\rescheduling\,\textbf{($P_2$)}};
  \node[align=center] at (0,-2.05) {Transportation\\operations\,\textbf{($P_4$)}};
  \node[align=center] at (0, 3.30) {Production\\rescheduling\,\textbf{($P_5$)}};
  \node[align=center, text=black!45] at (0,-3.30) {Production operations\\(out of scope)};
  % proximity arrow
  \draw[-{Latex[length=2mm]}, thick, draw=revblue!70] (4.4,3.4) to[out=-90,in=90] (4.4,0.2);
  \node[rotate=-90, text=revblue!80, font=\scriptsize] at (4.85,1.8) {proximity $\rightarrow$ priority};
\end{tikzpicture}
\caption{Domain-shell model. Concentric shells order application domains by structural proximity to the target (railway $\subset$ transportation, with production as an analogical outer shell); the horizontal split separates the target activity, rescheduling, from the wider set of operations. Badges $P_1$--$P_5$ give the corpus inclusion priority (activity-first, then domain); production operations is excluded.}
\label{fig:domain_shells}
\end{figure}

\subsubsection{Construction protocol}

Mining requires breadth: a taxonomy is only as defensible as the coverage of the corpus it is induced from. The protocol therefore admits formulations broadly within a fixed quality order and defers selectivity to the validation stage, where a single representative per cluster is examined in depth (Section~\ref{sec:validation}). This separation lets the corpus be large without diluting the rigor of the fidelity check.

\paragraph{Reporting standard.} The corpus construction follows the PRISMA guidelines for systematic reviews, reporting identification, screening, eligibility and inclusion with counts at each stage so the selection is auditable and reproducible.
\citep{page2021prisma}
Within the full corpus we additionally mark a \emph{core corpus} (the subset that satisfies the strict PRISMA eligibility criteria and the top quality tier) and report it separately, so that results can be read either over the broad mining corpus or over the smaller, fully review-grade core.

\paragraph{Quality tiers.} Sources are admitted in descending order of editorial scrutiny: (i) peer-reviewed journal articles and proceedings of established operations-research and transportation venues; (ii) other peer-reviewed conference papers and book chapters; and only to fill residual gaps in under-populated clusters, (iii) theses, technical reports and preprint servers such as arXiv. A formulation enters from a lower tier only when the higher tiers leave its cell or cluster below a documented minimum support. Predatory and pay-to-publish outlets whose editorial control is known to be weak (typically broad, generic-scope journals run on a high-throughput author-pays model) are excluded outright, irrespective of tier.

\paragraph{Source artifact.} Orthogonal to the venue tier is the form in which a formulation is published. A formulation accompanied by released solver code bound to its source paper is the highest-fidelity source: it can be ingested deterministically and re-solved within seconds, so it doubles as a validation anchor (Section~\ref{sec:validation}). Formulations available only as \LaTeX{} or prose are admitted as well and parsed accordingly (Section~\ref{sec:extraction}). Released code that is \emph{not} attached to a published formulation (unpublished or student repositories) is out of scope, since neither provenance nor peer scrutiny can be established for it.

\paragraph{Search and expansion.} Each priority cell $P_1$--$P_5$ is seeded with a keyword query (domain terms $\times$ optimization terms, e.g.\ ``train timetable rescheduling MILP''). The seed set is then expanded by systematic snowballing in both directions (backward over the reference lists and forward over the citing papers) until no new in-scope formulations appear. The search date is frozen and every query string is recorded so the corpus can be regenerated.

\paragraph{Unit of analysis and inclusion.} The unit is a \emph{formulation}: one optimization model comprising at least an objective and a constraint set that can be recovered from the paper's text, an appendix, or accompanying solver code. A single paper may contribute several formulations (e.g.\ a base model and an extension), each treated independently with a shared provenance record. A candidate is \emph{included} iff it (a) falls in an in-scope cell, (b) is an LP or MILP (a linear objective and linear constraints, with integer or binary variables for the MILP case), and (c) exposes a recoverable formulation. It is \emph{excluded} if it presents only a verbal or heuristic procedure with no explicit model, is nonlinear without a linearized variant, or duplicates an already-ingested formulation. These criteria are mutually exclusive and, over the candidate pool, collectively exhaustive.

\paragraph{Provenance.} Every ingested formulation carries a record: source identifier, venue and tier, year, citation count at the freeze date, domain shell, activity, and the priority cell $P_k$. Citation count is retained specifically to drive the \emph{a posteriori} selection of validation representatives (Section~\ref{sec:validation}); it is not used during clustering, so that popularity cannot bias the induced structure.

\paragraph{Reproducibility and threats.} The protocol is reproducible up to the freeze date: identical queries, tier thresholds and deduplication produce an identical corpus. Three threats are acknowledged and carried forward to the discussion of validity. Citation count is an imperfect proxy for importance and is recency-biased; venue quality is heterogeneous across sub-fields; and formulations that are well formalized in machine-readable form are easier to ingest, which can over-represent recent or code-released work. The current corpus comprises \prismaInclFormulations{} candidate formulations from \prismaInclPapers{} source papers (pre-review); Figure~\ref{fig:prisma} reports the full PRISMA flow, and the per-cell $P_1$--$P_5$ distribution follows from the domain/activity classification (in progress).

\begin{figure}[!t]
\centering
\begin{tikzpicture}[font=\scriptsize, node distance = 4mm,
  box/.style = {rectangle, rounded corners=2pt, draw=black!70, fill=black!4,
                text width=0.84\linewidth, align=left, inner sep=4pt, minimum height=7mm},
  fl/.style  = {-{Latex[length=2mm]}, thick, draw=black!70}]
\node[box] (id) {\textbf{Identification.} Database search (\prismaDBqueries{} frozen queries): \prismaDBrecords{} records; \prismaDBdups{} duplicates removed $\rightarrow$ \prismaDBunique{} unique. Other methods (citation searching / snowballing): \prismaCiteIdentified{} records identified.};
\node[box, below=of id] (sc) {\textbf{Screening.} \prismaDBunique{} database records screened against the in-scope cells $P_1$--$P_5$.};
\node[box, below=of sc] (el) {\textbf{Eligibility.} \prismaReportsRetrieved{} reports retrieved and assessed; \prismaReportsExcluded{} excluded (\prismaExclNotEntitled{} full text not entitled, \prismaExclNoFormulas{} no machine-readable formulas, \prismaExclTierThree{} awaiting Tier-3 PDF extraction).};
\node[box, below=of el] (in) {\textbf{Included.} \prismaInclPapers{} source papers contributing \prismaInclFormulations{} candidate formulations enter the mining pipeline.};
\draw[fl] (id) -- (sc); \draw[fl] (sc) -- (el); \draw[fl] (el) -- (in);
\end{tikzpicture}
\caption{PRISMA flow of corpus construction. Counts are generated deterministically from the corpus artifacts (\texttt{corpusbuilder.prisma}) and re-imported as macros, so the figure tracks the corpus exactly. Counts shown are the current pre-review snapshot.}
\label{fig:prisma}
\end{figure}

\subsection{The LP2Graph Representation}\label{sec:lp2graph}

Mining structure requires a representation that is precise, comparable across papers, and computable without solving. \emph{LP2Graph}\footnote{Open-source library \texttt{lp2graph}; see the repository for the canonical schema, view derivations and validation suite.} provides one. It represents an LP, MIP or MILP as a \emph{typed variable--equation graph} derived deterministically from a single canonical data model: identical input always yields byte-identical output, with no language model, sampling or temperature anywhere in the procedure. This determinism is what makes the mined dataset reproducible. The representation covers a \emph{defined} class of formulations, not all of mathematical programming; its scope, the distinction between what it can represent and what it can solve, and what the graph preserves, normalizes and loses are stated explicitly at the end of this section.

\subsubsection{Canonical model}

A formulation is a schema-versioned document declaring five parts. \emph{Index families} (e.g.\ trains, events, time slots) may be \emph{ordered} (admitting offsets such as $t-1$) and \emph{cyclic} (wrapping modulo their cardinality, as in periodic timetabling). \emph{Parameters} carry a \texttt{kind} (scalar, vector, matrix, big-M, tolerance); the big-M and tolerance tags are semantic, not decorative, and surface later as structural signals. \emph{Variable templates} declare a shape over index families, a domain (continuous, integer, binary) with bounds, and a \emph{role} (primary, auxiliary, slack, indicator); a template $y[I,I]$ stands for the whole family of $|I|^2$ variables. \emph{Constraint templates} are a comparison of two term lists under \emph{quantifiers} over index families, with restrictions ($i\neq j$, ordered pairs, attribute-based \texttt{where}-predicates). The \emph{objective} declares a sense and a combination rule (sum, weighted sum, lexicographic). Variables, parameters and constraints may additionally carry an \emph{additive semantic facet} (a domain-role for variables, a domain-class for parameters and constraints) recorded alongside the structural fields; it is optional and never required for validity, and is read into the type signature during homologization (Section~\ref{sec:extraction}). The atomic unit is a \emph{term}: a tuple of a referent (variable, parameter or literal), one binding per index slot (with offset and modulo), a coefficient, a sign, a role, and an optional aggregation operator ($\sum$, $\max$, $\min$, $\mathrm{abs}$, indicator, modulo) with the index families it ranges over. This single term grammar is what allows several graph views to be derived from one source of truth. Two-phase validation guards integrity: a JSON-schema check on document shape, then a semantic check on cross-cutting invariants: every referenced name is declared, every binding covers the referent's shape, and the family\footnote{In the library the model class is the \texttt{family} field, taking values \texttt{lp}/\texttt{mip}/\texttt{milp}; this LP/MIP/MILP class is distinct from the \emph{constraint families} induced at Level~C (Section~\ref{sec:feature_clustering}).} is consistent (an LP carries no integer or binary variable, an MILP at least one).

\subsubsection{Typed variable--equation graph and views}

The structural object is a typed variable--equation graph $G=(N,E)$. Nodes $N$ partition into classes: one per index family, one per parameter, one per variable template, one per constraint template, the objective, and one aggregation-operator node inserted between a container and the variables it aggregates. Edges $E$ are also typed: a \emph{var-in-constraint} or \emph{var-in-objective} edge for each term, an edge from each variable or parameter to the index families that shape it, and operator input/output edges. Restricting attention to the variable and equation (constraint/objective) classes recovers the familiar bipartite intuition (variables on one side, the equations that govern them on the other) while the typing retains everything needed for finer analysis.

Three deterministic view functions expose this object at different resolutions. The \emph{schema view} is the topology alone: templates and indices, with offsets suppressed. It is the canonical structural fingerprint (any two formulations with the same template skeleton produce the same schema graph) and is therefore the basis for cross-formulation comparison and for all topological metrics in this paper. The \emph{hybrid view} enriches the schema view with per-binding offset, sign and modulo labels, and is used for inspection. The \emph{ground view} materializes every variable and constraint instance at supplied index cardinalities and applies degeneracy filters (out-of-range offsets dropped, restricted tuples excluded, cyclic bindings wrapped); it is used for rendering at small sizes and, exported to PyG/DGL, as input to downstream graph learning. Export adapters to NetworkX, LaTeX and Pyomo are available for interoperability. Figure~\ref{fig:graph_example} illustrates the representation on a small example.

\begin{center}
\captionsetup{type=figure}
\begin{minipage}[c]{0.42\linewidth}
\centering
\footnotesize
\textbf{(a) A small formulation}
\begin{align*}
\min_{t,y}\ & \textstyle\sum_{i\in I} t_i \\
\text{s.t.}\ & t_j - t_i \ge h - M\,(1-y_{ij}),\\
 & \qquad \forall\, i,j\in I,\ i<j,\\
 & t_i \ge 0,\quad y_{ij}\in\{0,1\}.
\end{align*}
\end{minipage}\hfill
\begin{minipage}[c]{0.55\linewidth}
\centering
\footnotesize
\textbf{(b) Typed variable--equation graph}\\[6pt]
\begin{tikzpicture}[font=\scriptsize, node distance=5mm and 16mm,
  var/.style={rectangle, rounded corners=2pt, draw=black!70, fill=black!8, inner sep=3pt, minimum height=5mm},
  par/.style={rectangle, draw=black!55, fill=black!2, inner sep=3pt, minimum height=5mm},
  eq/.style ={rectangle, rounded corners=2pt, draw=revblue!70, fill=revblue!6, inner sep=3pt, minimum height=5mm},
  idx/.style={circle, draw=black!55, fill=black!12, inner sep=1.5pt},
  e/.style  ={draw=black!55}]
  \node[var] (t) {$t[I]$};
  \node[var, below=of t] (y) {$y[I,I]$};
  \node[par, below=of y] (h) {$h$};
  \node[par, below=of h] (M) {$M$};
  \node[idx, left=7mm of y] (I) {$I$};
  \node[eq, right=of t] (z) {$z$: $\min\sum t$};
  \node[eq, below=8mm of z] (c) {$C_1$: headway};
  % shape edges (index family -> variable templates)
  \draw[e] (I) -- (t);
  \draw[e] (I) -- (y);
  % var/param in equations
  \draw[e] (t) -- (z);
  \draw[e] (t) -- (c);
  \draw[e] (y) -- (c);
  \draw[e] (h) -- (c);
  \draw[e] (M) -- (c);
\end{tikzpicture}
\end{minipage}

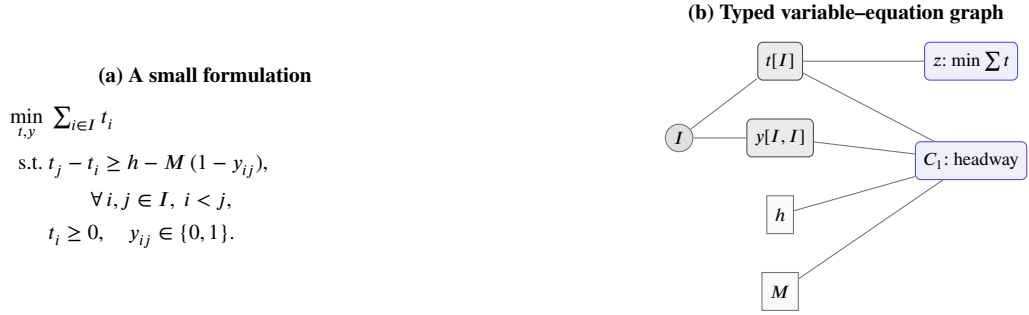
\captionof{figure}{The LP2Graph representation on a small big-M train-ordering model. (a) the formulation; (b) its typed variable--equation (schema) graph: variable templates ($t[I]$, $y[I,I]$, grey rounded), parameters ($h$, $M$, square), the index family $I$ (circle) that shapes the variables, and the constraint and objective nodes (blue) they connect to. The node typing is what lets structurally similar formulations be compared regardless of the symbols their authors chose.}
\label{fig:graph_example}
\end{center}

\subsubsection{Deterministic metrics}

Every metric is a pure function of the canonical model or its schema graph. Two \emph{well-formedness} indicators are used as structural features (Section~\ref{sec:feature_clustering}). \emph{Model coherence} is the indicator that the undirected schema graph is connected; it doubles as a data-quality signal during ingestion, since a disconnected schema graph flags an extraction in which a declared variable never enters any equation. \emph{Model completeness} is the indicator that a formulation declares an objective together with at least one variable and at least one constraint: the minimal evidence that a recoverable model, rather than a fragment, was extracted. The library additionally exposes a heuristic keyword-based constraint classifier over each constraint's name and description; we use it only as a seed for labeling (Section~\ref{sec:labeling}), never as ground truth.

The library further computes a family of structural \emph{complexity} metrics (presence flags $\boldsymbol{\phi}$; minimal size $S_{\min}=nV_{\min}\cdot nC_{\min}$; constraint--variable ratio $R_{C/V}=nC_{\min}/nV_{\min}$; graph diameter $D_G$; edge density $|E|/(|N|(|N|-1))$). They are not used as clustering features in this version; whether they add discriminative value beyond the type signatures is assessed with the corpus-scale ablations (Section~\ref{sec:feature_clustering}).

%% PARKED 2026-06-22: complexity metrics, pending a decision on whether they
%% earn their place in this paper. Restore this block (and their feature-vector
%% uses in sec_feature_clustering) if kept; see the visible note above.
% The first is a set of \emph{presence flags} $\boldsymbol{\phi}(f)\in\{0,1\}^{5}$ computed in one pass over the terms: the presence of a big-M construction, of integer or binary variables, of a modulo offset, of soft/slack variables, and of an aggregation operator. The second is a set of \emph{structural metrics} over the schema graph. Let $\mathcal{V}\subseteq N$ be the variable-template nodes and $\mathcal{C}\subseteq N$ the constraint-template nodes, with $nV_{\min}=|\mathcal{V}|$ and $nC_{\min}=|\mathcal{C}|$ counting indexed families once regardless of dimension. Minimal size measures model size,
% \begin{equation}
% S_{\min} = nV_{\min}\cdot nC_{\min},
% \label{eq:min_complexity}
% \end{equation}
% the constraint--variable ratio measures constraint density relative to the decision space,
% \begin{equation}
% R_{C/V} = \frac{nC_{\min}}{nV_{\min}},
% \label{eq:constraint_variable_ratio}
% \end{equation}
% and the graph diameter captures the longest dependency chain as the maximum shortest-path distance over the largest connected component of the undirected schema graph,
% \begin{equation}
% D_G = \max_{u,v\in N} d(u,v).
% \label{eq:graph_diameter}
% \end{equation}
% Edge density, $|E|/(|N|(|N|-1))$, was recorded alongside coherence as a further graph metric.

\subsubsection{Deterministic text\,$\leftrightarrow$\,graph codec and solving}

Beyond the views, LP2Graph offers a bidirectional, deterministic codec between a formulation's text and its graph, and a real solver back-end: the two capabilities the validation in Section~\ref{sec:validation} relies on. \texttt{to\_canonical\_latex} renders a paper-style \LaTeX{} document (calligraphic set notation $\mathcal{E}$, $\sum$, $\forall$, big-M), and \texttt{from\_canonical\_latex} parses it back, resolving each subscripted symbol through the declared symbol table. The codec is a tested fixed point: serialization is idempotent and the round-trip preserves the canonical normal form, with only incidental labels normalized; the solvable content (sets, parameters, variables, constraints, objective) round-trips exactly. The solver back-end grounds a formulation with an instance (index cardinalities plus parameter values) into a concrete linear program and solves it with CBC, HiGHS or Gurobi, so that the same model can be checked across independent solvers. Together these make a formulation not only comparable but \emph{executable} from its mined representation.

\subsubsection{Scope and boundary conditions}\label{sec:lp2graph_scope}

Because the paper's central claim is structural comparability, the limits of the representation must be as explicit as its capabilities. The canonical grammar admits linear formulations built from templates over regular index families: variables with domains continuous, non-negative, integer or binary and optional finite bounds; constraints comparing two linear term lists under $\le$, $\ge$ or $=$, quantified over index families with pairwise restrictions and attribute-based \texttt{where}-predicates; coefficients that are numeric literals or declared parameters; and single-level summations (no nested $\sum\sum$). Indicator constraints are a first-class construct (a constraint template may carry a gating binary and an activation value), and big-M formulations are explicit rather than implicit: big-M constants are tagged parameters, and a deterministic transform linearizes an indicator constraint by computing the tightest finite $M$ from the declared variable bounds. The objective is single but supports three combination rules: plain sum, weighted sum, and lexicographic priorities.

Within this class the library distinguishes what it can \emph{represent} from what it can \emph{solve}, and the distinction matters for which fidelity claims (Section~\ref{sec:validation}) a model can carry. Aggregation operators beyond summation ($\max$, $\min$, $\mathrm{abs}$, indicator terms, modulo terms) and lexicographic objectives are representable (they appear as typed operator nodes in the graph and count fully toward structural comparison), but the grounder rejects them for direct solving with an explicit \emph{unsupported-model} error: an indicator constraint must first be linearized by the big-M transform, a lexicographic objective solved as a sequence of single-objective problems, and the remaining operators await linearization rules. Outside the schema altogether are quadratic and other nonlinear terms, piecewise-linear constructs, SOS sets, semi-continuous variables, and solver-specific devices such as lazy constraints or callbacks; a source model using them is recorded with an explicit extraction status (Section~\ref{sec:extraction}) rather than approximated silently.

The graph \emph{preserves} template-level structure exactly: the typed nodes and edges, index shapes, quantifier patterns and restrictions, per-binding offsets and modulo wraps, domains, roles, and the semantic parameter and constraint tags. It \emph{normalizes} away author-specific surface form: symbol names (recovered separately by homologization, Section~\ref{sec:extraction}), notational variants of the same term, and term order, all mapped to a canonical normal form. It \emph{loses} whatever is not structural: concrete instance data (re-attached only at grounding time) and any prose semantics not captured by the optional semantic facets. On this basis \emph{structural equivalence} has a precise meaning: two formulations are structurally equivalent when their schema graphs are isomorphic respecting node and edge types (identical template skeletons up to renaming of symbols and index families), while finer distinctions (offsets, signs, modulo wraps) are compared on the hybrid view. Equivalence is thus a statement about model \emph{architecture}, deliberately insensitive to instance size and notation, and it is the invariant on which all cross-formulation comparison in this paper rests.

\subsection{Extraction and Homologization}\label{sec:extraction}

Every source formulation must become a validated LP2Graph canonical model before any analysis. Two facts make this non-trivial: papers express the same construct in incompatible notations, and the deterministic codec of Section~\ref{sec:lp2graph} parses only the canonical grammar, not arbitrary author notation. We therefore separate \emph{extraction} (recovering a canonical model from a heterogeneous source) from \emph{homologization} (reducing the recovered symbols to a common, comparable form).

\subsubsection{Extraction}

Extraction is the one stage of the pipeline that is \emph{not} claimed to be deterministic, and we state its boundary precisely: once a formulation has been extracted into the LP2Graph canonical model and certified by the two-phase validator, everything downstream (representation, feature construction, clustering, labeling and reporting) is deterministic; the path \emph{from source to canonical model} involves interpretation whose difficulty depends on the source channel. We therefore distinguish the channels explicitly, ordered by decreasing reliability.

\emph{(i) Solver code.} A formulation published as executable code (Pyomo, GAMS, AMPL, JuMP) is the preferred input: its sets, variables and constraints are already symbolic and machine-readable, so an importer maps them onto the canonical schema with essentially no interpretation. \emph{(ii) Canonical \LaTeX{}.} A formulation already expressed in, or mechanically transcribable to, the canonical grammar is parsed directly by the library's deterministic parser. \emph{(iii) Publisher full-text markup.} The bulk of the corpus arrives through structured publisher interfaces, full-text XML with MathML equations (Elsevier's text-and-data-mining API) and \LaTeX{} sources (arXiv), from which each display equation is recovered as machine-readable \LaTeX{}. This channel does \emph{not} operate on rendered PDFs; markup-to-\LaTeX{} conversion is rule-based, but the subsequent step, reconstructing the symbol table (which letters denote index families, parameters and variables, and their shapes) and rewriting each declaration and constraint into canonical templates and terms, requires reading the surrounding prose and is where interpretation enters. \emph{(iv) Manual transcription.} A formulation available only as typeset mathematics or prose is transcribed by hand into the canonical grammar. \emph{(v) Manual correction.} Any automatic result may be corrected by hand; corrections are first-class events, not silent overwrites.

Whatever the channel, the \emph{output} is always a canonical document re-checked by the same deterministic validator, and every transformation, automatic or human, is logged with provenance (channel, tool version, source locator, correction history) so each model traces back to its source lines and the human share of the reconstruction is reportable, not hidden. The human decisions of channels (iv) and (v), namely accepting, correcting or rejecting each recovered formulation and screening papers for scope, are collected through a dedicated mobile review instrument that presents each formulation with its source rendering and records every decision as a versioned, exportable event, so the manual share of the corpus is elicited systematically rather than ad hoc; a gamified variant of the same instrument opens this review to additional annotators.\footnote{A public demonstrator of the review instrument, restricted to a small openly licensed (CC-BY) subset of the corpus, is available at \url{https://railpmining.joernmaurischat.de/}.} Sources that resist extraction are themselves data: a paper whose model is documented incompletely, inconsistently or ambiguously is recorded as such (via the completeness and coherence indicators of Section~\ref{sec:lp2graph} and an explicit extraction-status field) rather than silently repaired, so the corpus also yields an honest picture of documentation practice in the field.

\paragraph{Extraction-quality audit.} Because the taxonomy can only be as reliable as the extraction feeding it, extraction quality is audited directly rather than assumed. A stratified sample of formulations, covering all source channels, is independently extracted a second time by a human into a gold canonical model, and the pipeline's model is compared against it element-wise: precision and recall over index families, parameters, variables and constraints (matched under the homologization equivalence of Eq.~\eqref{eq:homologization}), together with a categorization of every disagreement (missed element, spurious element, wrong domain or shape, wrong comparator or term, under-specified source). \vnext{report audit sample size, per-element precision/recall per channel, and the error-category distribution.}

\subsubsection{Homologization}

A variable named $x$ in one paper and $z$ in another may denote the same construct; conversely, the same letter may denote different things. Homologization removes this accidental variation so that entities are compared by what they \emph{are}, not by the symbol an author happened to choose. We define it as a map on each extracted symbol $s$ (a variable, parameter or constraint),
\begin{equation}
h(s) = \big(\tau(s),\ \kappa(s)\big),
\label{eq:homologization}
\end{equation}
with two independent components. The \emph{type signature} $\tau(s)$ is read directly from the canonical model: for a variable, its algebraic domain, structural role, the ordered list of index families of its shape, and, where the model assigns one, a semantic domain-role; for a parameter, its kind, shape and optional domain-class; for a constraint, its comparator, the quantifier and restriction structure, the multiset of referent kinds in its terms, and its declarative kind and optional domain-class. Type signatures are invariant to index renaming, so $x[I,T]$ and $z[A,B]$ with matching domains and family structure share a signature. The \emph{concept} $\kappa(s)$ is obtained by normalizing the symbol's name and its surrounding description (subscripts split, case folded, lemmatized, stop-words removed) and mapping the resulting meaningful tokens onto a concept identifier through the shared lexical resource of Section~\ref{sec:feature_clustering}; this is what collapses ``headway'', ``minimum separation'' and ``safety distance'' to one concept. Two entities are \emph{homologous} when they agree on both components, $s\sim s' \iff \tau(s)=\tau(s')\ \wedge\ \kappa(s)=\kappa(s')$, up to index renaming.

Homologization is applied bottom-up. Variables and parameters, the atomic elements of any LP, are homologized first, because their signatures and concepts are needed to characterize the constraints and objective that reference them (Section~\ref{sec:feature_clustering}). The procedure is deterministic given a fixed lexical resource and concept map; the precise normalization rules and the concept inventory are versioned, so re-running homologization on the same corpus reproduces the same homologized entities. Residual ambiguity (a symbol whose role or concept cannot be assigned automatically with sufficient confidence) is queued for human adjudication and the resolution is written back, with provenance, into the same versioned store used for labeling, closing the loop between extraction and the label database (Section~\ref{sec:labeling}). Extraction error is the dominant threat to this stage and is quantified directly in Section~\ref{sec:validation} by regenerating each validated model's \LaTeX{} and comparing it against the source.

\subsection{Feature Construction and Multi-level Clustering}\label{sec:feature_clustering}

The taxonomy is induced, not imposed. A single \emph{cluster-and-name} operator is defined once and then applied at every level of the model, each time over features appropriate to that level. Defining it once is what makes the procedure uniform and auditable.

\subsubsection{The cluster-and-name operator}

Let $E$ be a finite set of textual entities, each carrying a name and, optionally, a short description. The operator $\mathrm{CN}(E)$ returns a partition of $E$ together with a name for each part, in three deterministic steps.

\paragraph{(1) Meaningful tokens to concepts.} Each entity's text is tokenized, case-folded, split on subscripts and compounds, lemmatized, and reduced to content words by removing function words and a domain stop-list (so ``railway'' is kept and ``is'' discarded). A concept map $g$ sends each remaining token to a concept $c$ in a fixed vocabulary $\mathcal{C}$, grouping synonyms primarily through a frozen, versioned domain thesaurus (the deterministic backbone, covering domain terms that general resources miss, e.g.\ ``headway'', ``rolling stock'') and, optionally, through an acknowledged lexical database (WordNet synsets) \citep{miller1995wordnet,fellbaum1998wordnet}
when its pinned release is available; a token matched by neither falls back to its lemma. Grouping by frozen, versioned resources, rather than by ad-hoc judgment, is what keeps the concept assignment reproducible.

\paragraph{(2) Concept vector.} For entity $e$ with token multiset $T(e)$, the count of concept $c$ is $n_{e,c}=|\{w\in T(e): g(w)=c\}|$, weighted by a smoothed term frequency--inverse document frequency over the collection,
\begin{equation}
\hat{w}_{e,c} = n_{e,c}\;\cdot\;\Big(\ln\frac{1+|E|}{\,1+|\{e'\in E: n_{e',c}>0\}|\,}+1\Big),
\label{eq:tfidf}
\end{equation}
the additive smoothing (the $1+$ terms and the trailing $+1$) keeping the weight strictly positive and finite even for a concept that occurs in every entity or in none. The raw vector $\hat{\mathbf{v}}(e)=(\hat{w}_{e,c})_{c\in\mathcal{C}}$ is then $L^2$-normalized, $\mathbf{v}(e)=\hat{\mathbf{v}}(e)/\lVert\hat{\mathbf{v}}(e)\rVert_2$, so entities are compared under cosine distance and the per-entity token total cancels; common concepts that fail to discriminate are down-weighted automatically. At the structural levels below, the structural facets from LP2Graph are folded into the \emph{same} vector rather than kept as a separately scaled block: the categorical type-signature components enter directly as one-hot tokens, weighted by the scheme of Eq.~\eqref{eq:tfidf}. (The presence flags and the continuous complexity metrics are not used in this version; see Section~\ref{sec:lp2graph}.) The result is a single homogeneous vector $\tilde{\mathbf{v}}(e)$ on which no block dominates by scale.

\paragraph{(3) Cluster and name.} A clustering algorithm $\mathcal{A}$ partitions $\{\tilde{\mathbf{v}}(e)\}$ under cosine distance into clusters $C_1,\dots,C_K$ plus an explicit ``unassigned'' set, so that every entity lands in exactly one part. The default $\mathcal{A}$ is a deterministic agglomerative clustering (average linkage under cosine distance, with a distance threshold that discovers $K$ and leaves genuine outliers as singletons), chosen so that the partition is exactly reproducible and free of any external dependency; a fixed-$K$ variant selects $K$ by the silhouette criterion \citep{rousseeuw1987silhouettes},
and a density-based backend (HDBSCAN) \citep{campello2013hdbscan,mcinnes2017hdbscan}, which infers $K$ and isolates outliers rather than forcing them into clusters, is available as an option.
Each cluster is named after its most characteristic concepts (the ``most common formulation within'') by aggregating the normalized concept weights of Eq.~\eqref{eq:tfidf},
\begin{equation}
\ell_j = \operatorname*{arg\,max}_{c\in\mathcal{C}} \sum_{e\in C_j} v_{e,c},
\label{eq:naming}
\end{equation}
optionally reported as the top-$m$ concepts plus the medoid entity as an exemplar. The operator is thus $\mathrm{CN}(E)=\{(C_1,\ell_1),\dots,(C_K,\ell_K)\}$.

\subsubsection{Bottom-up structural levels}

The operator is applied in three passes, each consuming the output of the one below, so that higher-level structure is expressed in terms of lower-level vocabulary.

\paragraph{Level~V: variables and parameters.} Entities are the homologized variables and parameters. Features are the concept of the name, Eq.~\eqref{eq:tfidf}, concatenated with the type signature $\tau$ of Eq.~\eqref{eq:homologization} encoded categorically as one-hot tokens (algebraic domain, structural role, shape arity and index families, parameter kind, and the canonical model's semantic domain-role/domain-class facet where assigned). $\mathrm{CN}$ yields clusters of \emph{variable and parameter types} (e.g.\ binary sequencing indicators, continuous time/position variables, big-M parameters), each named by Eq.~\eqref{eq:naming}.

\paragraph{Level~C: constraints and objective.} Entities are the constraint templates and the objective. The feature vector conditions on Level~V: the concept of the constraint's name and description, concatenated with the histogram of Level-V clusters over the variables the constraint couples, and with the constraint's structural signature (comparator, quantifier/restriction pattern). Conditioning on which variable \emph{types} a constraint connects is what lets structurally identical couplings be recognized across different notations. $\mathrm{CN}$ yields \emph{constraint families} (and objective types), e.g.\ headway/separation, capacity, connection/transfer, flow conservation, big-M disjunctions.

\paragraph{Level~M: model type.} Entities are whole formulations. Features are the histogram of Level-C constraint families and Level-V variable types present in the model, concatenated with the model-level well-formedness indicators of Section~\ref{sec:lp2graph} (coherence and completeness). $\mathrm{CN}$ yields \emph{model types}: a structural answer to ``what kind of model is this'', derived from its parts rather than asserted by its authors.

\subsubsection{Text-only domain and approach}

Independently of the graph, two further labelings are induced from the paper text alone, because application domain and solution method are properties of the work, not of the formulation's structure. Running the operator's lexical pipeline (steps~1--2) over, respectively, the problem-description text and the methods text yields two one-dimensional clusterings: the application \emph{domain} (e.g.\ metro, mainline, freight) and the \emph{solution approach} (e.g.\ exact branch-and-cut, decomposition, rolling-horizon, matheuristic). Their product is a two-dimensional, text-only descriptor (domain $\times$ approach) attached to each model. Holding this descriptor apart from the structural levels is deliberate: it lets us later cross-tabulate \emph{what} is modeled (structure) against \emph{where} and \emph{how} it is solved (text), and to detect where authors' stated framing diverges from the structure they actually wrote.

\subsubsection{Determinism, cluster validity and ablations}

Given a frozen corpus, a versioned concept vocabulary $\mathcal{C}$, and fixed algorithm seeds, every pass reproduces the same partition and the same names. Cluster quality is reported with the silhouette coefficient; stability is reported as the adjusted Rand index across bootstrap resamples and across reasonable choices of $\mathcal{A}$ and of the vocabulary cutoff $|\mathcal{C}|$, so that a reviewer can see which clusters are robust and which are artefacts of a parameter choice. \vnext{report $K$ at each level, silhouette, and bootstrap ARI.}

An induced taxonomy invites a specific objection: that the clusters reflect naming conventions, notation, or corpus artefacts rather than mathematical structure. Because the feature vector deliberately mixes lexical and structural evidence, this is answered by ablation, not by assertion. The empirical evaluation therefore reports, at each level, the partitions obtained from \emph{lexical features only}, from \emph{structural features only} (type signatures, coupling histograms and structural signatures, with all concept tokens removed), and from the combined vector, together with the pairwise adjusted Rand index between the three: high lexical--structural agreement indicates that names and structure carry the same signal, while divergence localizes exactly where vocabulary and mathematics part ways. The same protocol covers the remaining design choices: clustering from variables only, constraints only, and the full model at Level~M; with and without homologization (raw symbols versus homologized entities), which quantifies how much comparability the normalization actually buys; and across the clustering backends and their parameters (agglomerative threshold, fixed-$K$ silhouette variant, HDBSCAN) and the feature weighting of Eq.~\eqref{eq:tfidf} versus unweighted counts. \vnext{ablation table, per level: lexical / structural / combined silhouette and pairwise ARI; with/without homologization; backend sensitivity.}

\subsection{Multi-level Labeling}\label{sec:labeling}

Clustering discovers structure but its parts are anonymous and corpus-specific. \emph{Labeling} assigns each entity a value from a controlled vocabulary that is stable, human-meaningful and reusable across corpora; the union of these vocabularies is the taxonomy. Labels are organized by level and dimension: at the variable level a \emph{type}; at the constraint level a \emph{family}; at the model level a structural \emph{type} together with the text-derived \emph{domain} and \emph{solution approach}. Each dimension has its own controlled vocabulary, seeded from the cluster names of Section~\ref{sec:feature_clustering} and then frozen and versioned.

\subsubsection{Two-stage mechanism with a closed loop}

Labels are assigned by a two-stage mechanism, deliberately combining the precision of explicit rules with the recall of a learned model, and feeding the learner's confident output back to grow the rules and the labeled set.

\emph{Stage~1: rule seed.} A transparent rule layer maps unambiguous evidence to labels: presence flags and type signatures from LP2Graph, and matches of an entity's concepts against a seed lexicon of domain-typical terms compiled from the highest-cited reference formulations. A rule returns a label or abstains, $r(e)\in\mathcal{Y}\cup\{\bot\}$. Rules are high-precision but low-recall; their value is a trustworthy labeled seed set $L=\{(e,r(e)):r(e)\neq\bot\}$ and an audit trail for every label they assign.

\emph{Stage~2: supervised generalization.} A classifier $f_\theta$ is trained on $L$ over the same feature vectors $\tilde{\mathbf{v}}(e)$ used for clustering, as a one-vs-rest model per class. Every dimension is treated as single-valued: the highest-scoring class is taken (variable type, constraint family, model type, and likewise the text-derived domain and solution approach); intrinsically multi-valued descriptors (a model spanning two domains or two solution approaches) are recognized but left to a planned extension and are not emitted by the current implementation. We default to a calibrated linear support-vector classifier (interpretable, effective on sparse TF-IDF features, and stable with the modest labeled sets typical here) whose calibrated probabilities gate write-back. The classifier labels the entities the rules left abstained.

\emph{Closed loop.} Predictions are not blindly trusted. A prediction with calibrated confidence above $\theta_{\text{high}}$ is accepted automatically; one below $\theta_{\text{low}}$, or one that contradicts a rule, is routed to a human adjudicator (active learning, which spends scarce human effort where it is most informative). Accepted and adjudicated labels are written back into the label database; newly reliable concept-to-label associations are promoted into the Stage-1 lexicon; and the classifier is retrained on the enlarged set. The loop repeats until few new labels are added. Algorithm~\ref{alg:labeling} states this precisely.

\begin{algorithm}[H]
\SetAlgoLined
\DontPrintSemicolon
\KwIn{entities $E$, features $\tilde{\mathbf v}$, rules $r$, thresholds $\theta_{\text{low}}\le\theta_{\text{high}}$}
\KwOut{versioned label database $D$, lexicon $\Lambda$, classifier $f_\theta$}
$L \leftarrow \{(e,r(e)) : e\in E,\ r(e)\neq\bot\}$\;
\Repeat{$|L|$ unchanged}{
  train $f_\theta$ on $L$\;
  \ForEach{$e\in E$ with no accepted label}{
    $(\hat y, p) \leftarrow f_\theta(e)$\;
    \uIf{$p \ge \theta_{\text{high}}$ \textbf{and} $\hat y$ consistent with $r(e)$}{accept $(e,\hat y)$ automatically\;}
    \uElseIf{$p < \theta_{\text{low}}$ \textbf{or} $\hat y$ conflicts with $r(e)$}{adjudicate $(e,\cdot)$ by a human\;}
    \Else{defer to next iteration\;}
    write accepted/adjudicated label to $D$ with provenance\;
    promote confirmed concept$\to$label rules into $\Lambda$\;
  }
  $L \leftarrow$ all confirmed labels in $D$\;
}
\caption{Two-stage labeling with closed-loop database update}
\label{alg:labeling}
\end{algorithm}

\subsubsection{Provenance, reproducibility and guardrails}

Every label record stores the entity, level, dimension, value, source (rule, classifier or human), confidence, and the versions of the lexicon, classifier and corpus in force when it was written. The rule stage is deterministic; the classifier is deterministic given a fixed seed and data version; the human decisions are logged, so the entire loop is reproducible by replaying the versioned decision log rather than by re-eliciting judgments.

Self-training that writes its own predictions back can amplify errors, so the design is explicitly guarded. The confidence gate and rule-consistency check bound what may be auto-accepted; a held-out, human-labeled gold set is re-scored after every loop to detect precision regressions and label drift, and the versioned store allows roll-back to any prior state. Subjectivity in the gold labels is quantified by inter-annotator agreement (Cohen's or Fleiss' $\kappa$) \citep{cohen1960kappa,fleiss1971kappa},
and class imbalance is surfaced by reporting per-class precision and recall rather than overall accuracy. The evaluation additionally reports how the labels were arrived at (the share assigned by rules, auto-accepted from the classifier, and human-adjudicated) and a qualitative error analysis: concrete examples of wrong or unstable labels (entities whose label changes across loop iterations or parameter choices), since these localize where the vocabulary is still ill-defined. \vnext{report seed-set size $|L|$, gold-set size and agreement $\kappa$, final per-dimension precision/recall, rule/auto/human label shares, and wrong- or unstable-label examples.} These choices keep the learned component from quietly corrupting a taxonomy whose chief value is its objectivity.

\subsection{Representation-Fidelity Validation}\label{sec:validation}

A taxonomy mined through LP2Graph is only trustworthy if the representation it rests on faithfully captures the source models. We therefore validate the representation explicitly rather than assume it, and we separate three fidelity claims of increasing strength so that each can be reported honestly.

\subsubsection{Three fidelity claims}

\emph{Structural fidelity} asks whether the typed graph preserves the model's structure. It is established by the deterministic codec of Section~\ref{sec:lp2graph}: serializing a canonical model to \LaTeX{} and parsing it back is idempotent and preserves the canonical normal form, $\textsf{parse}(\textsf{render}(f)) \equiv f$, with the solvable content (sets, parameters, variables, constraints, objective) round-tripping exactly. This holds for every model the canonical grammar admits and requires no instance data.

\emph{Translation fidelity} asks whether the canonical model still denotes the same optimization problem once expressed in an independent language. For a model with recoverable instance data, LP2Graph regenerates it both as paper-style \LaTeX{} and as an executable model in a second modeling language, grounds it with the instance, and solves it. Agreement between solving the regenerated model and solving the original confirms that nothing semantic was lost in mining.

\emph{External fidelity} asks whether the mined model reproduces the source paper's own result. The grounded model is solved with three independent solvers and checked against the published optimum,
\begin{equation}
z^{\star}_{\text{CBC}} = z^{\star}_{\text{HiGHS}} = z^{\star}_{\text{Gurobi}}
\quad\text{and}\quad
\bigl|\,z^{\star}_{\text{mined}} - z^{\star}_{\text{paper}}\,\bigr| \le \varepsilon .
\label{eq:external_fidelity}
\end{equation}
Cross-solver agreement rules out solver-specific artefacts; agreement with $z^{\star}_{\text{paper}}$ anchors the mined model to an externally established value. The whole pipeline (render, parse, ground, solve) is deterministic and free of any language model.

\subsubsection{Corpus-wide coverage and citation-anchored selection}

The three claims are applied at very different scales, and stating the scale of each is part of the validation design. \emph{Structural fidelity is checked for every canonical model in the corpus}: the codec round-trip needs no instance data and costs milliseconds, so a corpus-wide round-trip pass rate is reported, not sampled. \emph{Translation fidelity is attempted for every model the grammar admits as executable}: regeneration into the second modeling language is likewise data-free, and grounding is attempted wherever instance data are recoverable; the report states, for the whole corpus, how many models could be translated, how many grounded, and why the rest could not. Only \emph{external fidelity} is selective, because it needs both instance data and a published optimum, neither universally available. We apply it to one representative per cluster: the \emph{highest-cited} formulation in each label class, taken from the provenance record (Section~\ref{sec:scope_corpus}). Highest-citation representatives are the most scrutinized, the most influential, and the most likely to report reproducible numerics, so anchoring the taxonomy at these points validates its load-bearing members. Where the top representative lacks recoverable instance data, the next-highest is tried; where none in a cluster publishes usable data, a benchmark instance of the same structural class is substituted (for example a MIPLIB cyclic-timetabling instance for a PESP cluster) \citep{gleixner2021miplib},
and the cluster is marked as carrying structural and translation fidelity but not paper-anchored external fidelity.

Validation status is recorded per model and aggregated per cluster, so the dataset (Section~\ref{sec:outputs}) carries a \emph{validation-coverage table}: for each cluster, the fraction of members passing the round-trip, the fraction translated and grounded, and which fidelity level its anchor reached. Failures are data, not embarrassments, and every failure is categorized by cause: extraction error (the canonical model misrepresents the source), missing instance data, construct outside the grammar or unsupported by the grounder (Section~\ref{sec:lp2graph_scope}), cross-solver disagreement, or a source paper too ambiguous or under-specified to decide. The categories point to different remedies, and reproducibility failures in the published record are themselves a finding of the mining. \vnext{report corpus-wide round-trip rate, translation/grounding rates with reasons, per-cluster coverage table, and the failure-category distribution.}

\subsubsection{Coverage limits and threats to validity}

The canonical grammar covers template-structured models (periodic-event scheduling, assignment/matching, time-indexed packing and covering, big-M disjunctive ordering), and for these the pipeline reproduces independently verified optima exactly. Models outside it are reported, not hidden: formulations with per-train variable-length routes or a derived ``next resource'' cannot yet be expressed by the regular-index-family templates, so they receive structural labels where partially recoverable but are excluded from external-fidelity claims, and grammar extension is left as future work. Four threats are carried to the discussion. Instance data are often unpublished, capping how much of the corpus can reach external fidelity; published optima can themselves be wrong or under-specified; a single representative may not speak for its whole cluster; and grammar coverage is incomplete. The third threat is mitigated by reporting intra-cluster structural homogeneity (the rate of schema-graph isomorphism within each cluster) so a reviewer can judge how representative the validated anchor is. \vnext{report the fraction of clusters reaching each fidelity level, the tolerance $\varepsilon$, and intra-cluster isomorphism rates.}

\subsection{Outputs: Dataset and Taxonomy}\label{sec:outputs}

The method yields two artefacts. The first is a \emph{dataset}: for every formulation, its LP2Graph canonical model, schema and hybrid graphs, structural metrics and presence flags, homologized variables, parameters and constraints, multi-level and multi-dimensional labels, provenance, and validation status. The dataset is machine-readable and versioned, and it is \emph{regenerable}: given the frozen corpus, the versioned concept vocabulary and lexicon, and the recorded configuration, re-running the pipeline reproduces it identically, with any human adjudications replayed from the logged decision history rather than re-elicited.

The second artefact is the \emph{taxonomy}: the controlled vocabularies the labeling settles on, each entry defined by its characterizing concepts (Eq.~\eqref{eq:naming}) and illustrated by its medoid exemplar and corpus frequency. Table~\ref{tab:taxonomy_axes} lists its axes. The taxonomy is multi-level (variable, constraint, model) and multi-dimensional (structure-derived type; text-derived domain and solution approach).

\begin{table}[h]
\centering
\caption{Axes of the induced taxonomy. Categories are placeholders pending the empirical run.}
\label{tab:taxonomy_axes}
\small
\begin{tabularx}{\linewidth}{@{}l l l X@{}}
\toprule
\textbf{Level} & \textbf{Dimension} & \textbf{Derived from} & \textbf{Example categories} \\
\midrule
Variable/parameter & type & structure $+$ lexical & sequencing indicator; time/position variable; big-M parameter; slack \\
Constraint & family & structure $+$ lexical & headway/separation; capacity; connection/transfer; flow conservation; big-M disjunction \\
Model & type & structure & periodic-event (PESP); assignment/matching; time-indexed packing; big-M ordering \\
Model & domain & text & metro; mainline; freight; multimodal \\
Model & solution approach & text & exact branch-and-cut; decomposition; rolling horizon; matheuristic \\
\bottomrule
\end{tabularx}
\end{table}

The taxonomy is MECE by construction along well-defined boundaries. The dimensions are orthogonal (a model has a structural type \emph{and} a domain \emph{and} an approach independently), so an entity carrying several labels does so across dimensions, never as competing values within one. Within a single dimension the categories are mutually exclusive, and the current implementation assigns exactly one value per dimension (intrinsically multi-valued descriptors such as a multimodal domain are left to a planned extension), while exhaustiveness is guaranteed by a residual ``unassigned'' category that holds whatever the current vocabulary does not yet cover and that is emptied over successive refinement passes. What the method guarantees is objectivity and repeatability: the same inputs yield the same taxonomy, and every category traces to explicit features and named exemplars. What it does not guarantee is uniqueness (a different lexical resource or clustering choice could induce a different, equally valid partition), which is why cluster stability (Section~\ref{sec:feature_clustering}) and fidelity coverage (Section~\ref{sec:validation}) are reported alongside the taxonomy rather than asserted away.

\subsubsection{Release plan}

Because the dataset is a primary contribution, its release is specified concretely rather than promised in passing. \emph{What is released:} for every included formulation, the LP2Graph canonical model (JSON), its schema and hybrid graphs, homologized entities, feature vectors, labels, validation status, and a provenance record linking it to the source paper's DOI with the exact source locator, extraction channel and tool versions of Section~\ref{sec:extraction}; alongside these, the versioned taxonomy itself, the frozen concept vocabulary and seed lexicon, the logged human decisions (review verdicts, corrections and label adjudications), the PRISMA screening record of Section~\ref{sec:scope_corpus}, and the complete pipeline code (extraction, homologization, clustering, labeling and validation). \emph{What is not released:} the publishers' original full texts and equation markup. The corpus is mined from publisher interfaces under text-and-data-mining terms that permit publishing derived structural data but not redistributing the source text; the release therefore contains our derived representations and pointers (DOI, section, equation number) into the originals, which is also why every derived artefact is DOI-linked. \emph{Reproducibility:} a user with their own lawful access to the same interfaces can re-run the pipeline from the frozen query set and screening record to regenerate the corpus, and any user can regenerate every downstream artefact (graphs, features, clusters, labels, validation) from the released canonical models alone, since the pipeline is deterministic from that point on (Section~\ref{sec:extraction}). The taxonomy and label database are versioned, so later corpus growth extends rather than silently rewrites them. The library is already public under an open-source license; the dataset is released with a documented schema and, for downstream machine-learning use, a fixed documented split. \vnext{archive repository and DOI, dataset license, and split definition to be fixed at submission.}

\subsubsection{Further steps}

The dataset and taxonomy are a substrate, not an endpoint. Three uses follow directly. They enable \emph{retrieval}: given a new textual rescheduling problem, the structurally closest model types and constraint families can be looked up and reused. They enable \emph{structure-aware generation and validation}: an automated model builder can be conditioned on the taxonomy's categories rather than free-form text, and its output can be validated (checked structurally against the known model families before it is ever solved), turning model development into selection and recombination over a known structural space; this is the role our earlier \emph{raiLPminer} generation experiments attempted without such a substrate, and the foundation its successor builds on. And they enable \emph{learning}: the ground views export to graph-learning frameworks, so the labeled corpus can train models that predict type, family or solver-relevant structure. In each case the contribution of this paper (an objective, repeatable, validated structural inventory) is the precondition that makes the downstream automation scientifically defensible.

\section{Experiments}\label{sec:Experiments}

This section specifies how the mining pipeline of Section~\ref{sec:3} is applied
end-to-end to the railway-rescheduling corpus, and what each stage of the run
reports. The run answers three questions: which variable roles, constraint
families and model types recur across the published record; how the induced
structure compares to the vocabulary-based groupings of narrative surveys; and
how reproducibly the whole result can be regenerated from the released
artifacts. \emph{In this preliminary version the protocol is fixed and the
pipeline is demonstrated end-to-end on the library's bundled fixture set
(Section~\ref{sec:result}); the corpus-scale numbers follow in a subsequent
version once the human review of the extracted corpus is complete.}

\paragraph{Corpus.} The corpus is assembled under the domain-shell
prioritization and PRISMA protocol of Section~\ref{sec:scope_corpus}. At the
current freeze the identification and screening stages are complete:
\prismaDBqueries{} frozen database queries plus systematic two-directional
snowballing yield \prismaInclPapers{} source papers contributing
\prismaInclFormulations{} candidate formulations (Figure~\ref{fig:prisma}).
These candidates are machine-extracted but not yet human-reviewed; the
review--ingest step that promotes accepted formulations to canonical models is
the gate between this snapshot and the corpus-scale run.

\paragraph{Extraction and homologization.} Each candidate is parsed through the
channel hierarchy of Section~\ref{sec:extraction} (solver code where
released, publisher full-text markup (MathML) as the bulk channel, \LaTeX{} or
PDF otherwise) into the LP2Graph canonical model, then homologized under
Eq.~\eqref{eq:homologization}. The run reports per-channel coverage (candidates
parsed, admitted by the grammar, rejected with cause) and the
extraction-quality audit of Section~\ref{sec:extraction} over a stratified
sample spanning all channels.

\paragraph{Clustering and labeling.} The cluster-and-name operator
(Section~\ref{sec:feature_clustering}) runs with its deterministic default:
agglomerative clustering under cosine distance with a versioned concept
vocabulary and frozen thesaurus, sweeping variables, then
constraints/objective, then whole-model structure, plus the two text-only
dimensions. Labels are assigned by the two-stage mechanism of
Section~\ref{sec:labeling}. All resource versions (clustering, vocabulary,
rewrite rules) are pinned and reported, so the partition is exactly
reproducible; the ablation protocol of
Section~\ref{sec:feature_clustering} (lexical-only vs.\ structural-only vs.\
combined, with/without homologization, backend sensitivity) runs on the same
frozen inputs.

\paragraph{Validation.} Validation follows Section~\ref{sec:validation} at its
three scales: the codec round-trip over every canonical model in the corpus;
translation and grounding wherever the grammar admits execution; and external
fidelity for the highest-cited representative of each cluster, re-solved across
CBC, HiGHS and Gurobi against the optimum reported in the source paper, with
benchmark substitutes where no member of a cluster publishes usable instance
data. Outcomes are aggregated into the per-cluster validation-coverage table
and failure-category distribution released with the dataset
(Section~\ref{sec:outputs}).

\section{Results}\label{sec:result}

% =====================================================================
% PROVISIONAL DRAFT: TEST-SET RESULTS, NOT THE FINAL CORPUS.
% Everything in \prov{orange} below was generated by running the M3
% taxonomy induction (lp2graph clustering version cluster-2026.06.0) over
% the 10 canonical formulation fixtures shipped with the library
% (lp2graph/formulations/*.json), NOT over the mined railway corpus,
% which is still under construction. Regenerate over the final corpus and
% remove the orange flag once the corpus freezes.
% Reproduce with:  PYTHONPATH=src python3 -c "from pathlib import Path;
%   from lp2graph import load; from lp2graph.mining.cluster import induce;
%   print(induce([load(p) for p in sorted(Path('formulations').rglob('*.json'))]).summary())"
% =====================================================================

\prov{\noindent\textbf{[Preliminary: end-to-end demonstration, not the final
corpus.]} The figures and taxonomy in this section were produced by running the
M3 induction (\texttt{lp2graph} clustering version \texttt{cluster-2026.06.0})
over the \textbf{10 canonical formulation fixtures} bundled with the library,
\emph{not} over the mined railway corpus, whose human review is still in
progress (Section~\ref{sec:Experiments}). They demonstrate the pipeline
end-to-end on real, deterministically reproducible output; the corpus-scale
results will replace this section verbatim in a subsequent version. The cluster
names are the raw M3 cluster names (the M4 labeling layer, which maps them to
the controlled taxonomy vocabulary, is not applied here).}

\subsection{The mined dataset}

\prov{The test set comprises $N=10$ formulations (6 MILP and 4 LP) spanning
fixed-sequence dispatching, big-M disjunctive ordering, time-indexed packing,
periodic-event scheduling (PESP), assignment, multi-objective lateness/energy,
and absolute-deviation and lexicographic objectives. Because these are synthetic
fixtures rather than extracted models, no source-paper count ($M$) or citation
ranking is reported here; both arrive with the corpus.}

\subsection{The induced taxonomy}

\prov{The bottom-up induction yields $V=13$ variable/parameter-role clusters at
Level~V, $K=9$ constraint-family (plus objective) clusters at Level~C, and
$T=4$ model-type clusters at Level~M; the two text-only dimensions yield 9
domain and 9 solution-approach clusters. Mapping to the abstract's coverage
placeholder: $N=10$ formulations yield $K=9$ constraint families, $V=13$
variable roles and $T=4$ model types. Table~\ref{tab:taxonomy-testset} gives the
full level\,$\times$\,cluster\,$\times$\,count breakdown.}

\begin{table}[t]
\centering
\caption{\prov{\textbf{Preliminary} induced taxonomy over the 10-formulation
demonstration set (lp2graph \texttt{cluster-2026.06.0}). Each row is one M3
cluster with its member count $n$; names are raw M3 cluster names pending M4
labeling. To be regenerated over the final corpus in a subsequent version.}}
\label{tab:taxonomy-testset}
\prov{%
\begin{tabular}{llr}
\toprule
Level (entities) & Cluster (raw M3 name) & $n$ \\
\midrule
\multicolumn{3}{l}{\textbf{Level V: variable/parameter roles} (13 clusters, 37 entities)} \\
 & \texttt{kind:non\_negative} & 8 \\
 & \texttt{over:E}            & 6 \\
 & \texttt{kind:scalar}       & 4 \\
 & \texttt{target}            & 4 \\
 & \texttt{train}             & 3 \\
 & \texttt{assign}            & 2 \\
 & \texttt{energy}            & 2 \\
 & \texttt{indicator}         & 2 \\
 & \texttt{periodic}          & 2 \\
 & \multicolumn{1}{l}{\emph{4 singletons}: \texttt{bound, cap, crui, earliest}} & 1 each \\
\midrule
\multicolumn{3}{l}{\textbf{Level C: constraint families $+$ objective} (9 clusters, 24 entities)} \\
 & \texttt{kind:min/sum}      & 5 \\
 & \texttt{one}               & 4 \\
 & \texttt{over:E}            & 4 \\
 & \texttt{ref:variable}      & 4 \\
 & \texttt{cost}              & 3 \\
 & \multicolumn{1}{l}{\emph{4 singletons}: \texttt{choo, earliest, kind:min/lexicographic, slot}} & 1 each \\
\midrule
\multicolumn{3}{l}{\textbf{Level M: model types} (4 clusters, 10 entities)} \\
 & \texttt{ckind:linear}      & 3 \\
 & \texttt{vtype:target}      & 3 \\
 & \texttt{ckind:set\_packing}& 2 \\
 & \texttt{vtype:over:E}      & 2 \\
\midrule
\multicolumn{3}{l}{\textbf{Domain} (text-only; 9 clusters, 10 entities)} \\
 & \texttt{periodic}          & 2 \\
 & \multicolumn{1}{l}{\emph{8 singletons}: \texttt{abs, assignment, big, cost, fix, index, lexicographic, soft}} & 1 each \\
\midrule
\multicolumn{3}{l}{\textbf{Solution approach} (text-only; 9 clusters, 10 entities)} \\
 & \texttt{periodic}          & 2 \\
 & \multicolumn{1}{l}{\emph{8 singletons}: \texttt{abs, assignment, big, cost, fix, lexicographic, slot, soft}} & 1 each \\
\bottomrule
\end{tabular}%
}
\end{table}

\prov{The induction is deterministic: identical inputs and clustering version
yield byte-identical clusters. On this tiny test set many clusters are
singletons (expected, since 10 hand-built fixtures share little vocabulary);
the final corpus is where the cluster structure becomes informative.}

% =====================================================================
% TODO outline for the FINAL-CORPUS write-up (mining frame), to replace
% the provisional draft above once the corpus is built:
% - The mined dataset: number of formulations and source papers, per the
%   placeholder in the abstract ([PH: N formulations, M papers, K constraint
%   families, V variable roles, T model types]).
% - The induced taxonomy: the multi-level, multi-dimensional clusters and their
%   M4 labels (variable roles, constraint families, model types, plus the
%   text-only domain and solution-approach dimensions).
% - Structural metrics across the corpus: distributions of the deterministic
%   LP2Graph metrics (minimal complexity, constraint-variable ratio, graph
%   diameter) over the mined formulations.
% - Validation outcomes: round-trip fidelity, second-language agreement and
%   cross-solver optimum reproduction for the per-cluster representatives;
%   where structural fidelity holds and where semantic/optimality fidelity is
%   bounded by extraction accuracy and grammar coverage.

\section{Conclusion}\label{sec:concl}

LP Mining with LP2Graph turns dispersed, incompatibly-notated published LP and
MILP formulations into a reproducible dataset and an objective, structural
taxonomy of variables, constraints and model types. The method rests on a
single canonical model per formulation: once a source is extracted into it
(the one audited, non-deterministic step), everything downstream is
deterministic, from the typed variable--equation graph through homologization,
the bottom-up cluster-and-name sweep, and the two-stage labeling, so the entire
result regenerates by replay rather than re-judgment.

This preliminary version contributes the method itself: the canonical
representation and its scope conditions (Section~\ref{sec:lp2graph}), the
PRISMA-governed corpus protocol with \prismaInclPapers{} source papers and
\prismaInclFormulations{} machine-extracted candidate formulations already
screened (Section~\ref{sec:scope_corpus}), the full experimental and validation
protocol (Section~\ref{sec:Experiments}), and an end-to-end demonstration on
the library's fixture set showing that the pipeline runs deterministically from
canonical models to a named multi-level taxonomy
(Section~\ref{sec:result}). The corpus-scale taxonomy and validation outcomes
follow in a subsequent version once the human review of the extracted corpus is
complete.

The claims are deliberately bounded. Validation establishes \emph{structural}
fidelity corpus-wide via the codec round-trip, while semantic and optimality
fidelity are established only where translation, grounding and a published
optimum permit, and remain bounded by extraction accuracy and the coverage of
the canonical grammar. The method further depends on well-structured source
formulations and on the reach of the homologization resources: models outside
the regular-index-family templates are labeled where partially recoverable but
excluded from external-fidelity claims, and corpus coverage is itself limited
by publisher access and the freeze date.

The resulting taxonomy is intended as infrastructure: an objective map of which
modeling structures the field actually uses, on which the automated development
of railway-rescheduling optimization models (the generation line of work
this paper's validation and taxonomy feed) can be built.

%% --- Appendix ---
\appendix
% SKELETON: the previous appendix held the raiLPminer generation figures
% (full acceptance-rate analysis, constraint analysis by workflow/model and by
% paper, full binary-metric regressions, domain-oriented constraint coverage).
% Those belong to the generation line of work (Paper 2 / raiLParchitect) and have
% been removed; their PDFs live under appendix/. Any appendix material for Paper 1
% (LP Mining with LP2Graph), e.g. the full mined dataset, per-cluster
% representatives, or extended validation tables, is to be added here.

\section*{Declaration of Generative AI and AI-assisted technologies in the writing process}
During the preparation of this manuscript the authors used Claude (Anthropic), ChatGPT (OpenAI) and you.com (Claude, OpenAI, Meta, DeepSeek) to implement concepts into code and document it, support literature research, improve text writing style and translate results (e.~g.\ tables) into \LaTeX{}. All AI-generated text was reviewed and edited by the authors, and the authors accept full responsibility for the content and any errors.

\section*{Declaration of Competing Interests}
J\"orn Maurischat reports a relationship with DB Engineering und Consulting GmbH that includes: employment. All authors declare no other competing financial interests or personal relationships that could have appeared to influence the work reported in this paper.

%% --- Back matter ---
\printcredits

\bibliographystyle{cas-model2-names}
\bibliography{cas-refs}

\end{document}